\documentclass[conference]{IEEEtran}
\usepackage{amsmath,amsfonts}
\usepackage{algorithmic}
\usepackage{algorithm}
\usepackage{array}
\usepackage[caption=false,font=normalsize,labelfont=sf,textfont=sf]{subfig}
\usepackage{textcomp}
\usepackage{url}
\usepackage{verbatim}
\usepackage{graphicx}
\usepackage{cite}
\usepackage{booktabs}
\usepackage{multirow}
\usepackage{orcidlink}
\usepackage{xcolor} 
\usepackage{graphicx} 
\usepackage{tcolorbox}
\def\BibTeX{{\rm B\kern-.05em{\sc i\kern-.025em b}\kern-.08em
    T\kern-.1667em\lower.7ex\hbox{E}\kern-.125emX}}
\newcommand{\paragraphb}[1]{\vspace{-0.00in}\noindent{\bf #1} }

\begin{document}

\title{Decoding Neighborhood Environments with Large Language Models}

\author{
    \IEEEauthorblockN{
        Andrew Cart\IEEEauthorrefmark{1},
        Shaohu Zhang\IEEEauthorrefmark{1},
        Melanie Escue\IEEEauthorrefmark{1},
        Xugui Zhou\IEEEauthorrefmark{2},\\
        Haitao Zhao\IEEEauthorrefmark{1},
        Prashanth BusiReddyGari\IEEEauthorrefmark{1},
        Beiyu Lin\IEEEauthorrefmark{3},
        Shuang Li\IEEEauthorrefmark{4}
    }
    \IEEEauthorblockA{\IEEEauthorrefmark{1}University of North Carolina at Pembroke, USA}
    \IEEEauthorblockA{\IEEEauthorrefmark{2}Louisiana State University, USA }
    \IEEEauthorblockA{\IEEEauthorrefmark{3}University of Oklahoma, USA}
    \IEEEauthorblockA{\IEEEauthorrefmark{4}North Carolina A\&T State University, USA }
}

\maketitle

\begin{abstract}
Neighborhood environments include physical and environmental conditions such as housing quality, roads, and sidewalks, which significantly influence human health and well-being. Traditional methods for assessing these environments, including field surveys and geographic information systems (GIS), are resource-intensive and challenging to evaluate neighborhood environments at scale. Although machine learning offers potential for automated analysis, the laborious process of labeling training data and the lack of accessible models hinder scalability. This study explores the feasibility of large language models (LLMs) such as ChatGPT and Gemini as tools for decoding neighborhood environments (e.g., sidewalk and powerline) at scale. We train a robust YOLOv11-based model, which achieves an average accuracy of 99.13\% in detecting six environmental indicators, including streetlight, sidewalk, powerline, apartment, single-lane road, and multilane road.  We then evaluate four LLMs, including ChatGPT, Gemini, Claude, and Grok,  to assess their feasibility, robustness, and limitations in identifying these indicators, with a focus on the impact of prompting strategies and fine-tuning. We apply majority voting with the top three LLMs to achieve over 88\% accuracy, which demonstrates LLMs could be a useful tool to decode the neighborhood environment without any training effort.

\end{abstract}


\section{Introduction}
\IEEEPARstart{N}{eighborhood} environments refer to the community where people live and participate in daily life, including its physical and environmental conditions, which play a critical role in shaping human health, behavior, and quality of life \cite{ulijaszek2018physical, javanmardi2019chronic, nguyen2021county}. Those environmental indicators include housing quality, streetlights, parks, sidewalks, green space, power lines, etc. Research studies have shown the impact of neighborhood environments on health outcomes (e.g., obesity, diabetes, and mortality rates) \cite{larson2009food, phan2020gsv} and well-being factors (e.g., physical activity and access to nutritious foods) \cite{nguyen2022google, larson2009food}. For example, the presence of visible power lines has been associated with a higher prevalence of obesity and diabetes \cite{phan2020gsv}, while areas that were zoned for mixed use have been associated with lower health problems~\cite{nguyen2022google}. Meanwhile, research has also shown that the presence of other environmental indicators, such as green space, sidewalks, and parks, in an environment has increased its well-being \cite{giles2002relative,phan2020gsv, nguyen2022google}. For instance, the presence of greenspace is associated with fewer chronic diseases such as diabetes and obesity \cite{ nguyen2022google}. 

Traditionally, researchers have relied on field surveys \cite{schmidt2014survey}, remote sensing data \cite{kang2020remote}, and geographic information systems (GIS) \cite{pikora2002developing} to assess neighborhood characteristics. However, these methods often require extensive resources and may not capture the dynamic and nuanced aspects of human interactions within these environments. Google Street View (GSV) provides its application programming interface (API) to download street view images, enabling to use machine learning (ML) and deep learning (DL) to identify neighborhood environments in large geographic regions or large-scale data \cite{phan2020gsv, alirezaei2023multi }. However, it is laborious to label the images for training and validation, as no trained or well-tested models are available for detecting neighborhood environments.
    
Recent advances in artificial intelligence (AI), particularly large language models (LLMs), such as ChatGPT \cite{OpenAI_Chat_API}, DeepSeek \cite{deepseek}, and Gemini \cite{gemini} have shown promising capabilities in understanding and interpreting images \cite{ren2023chatgpt, zhang2024good}. LLMs train on vast amounts of image data, providing a potential method to analyze various sources such as transportation signs, store logos, sidewalks, and streetlights about local communities on a large scale. This gives us a new approach to identify the neighborhood environments without any model training effort. This raises our research questions.
\begin{itemize}
    \item[] \textbf{RQ1.} Can LLMs decode neighborhood environments accurately compared to supervised learning? 
    \item[] \textbf{RQ2.} What challenges and opportunities can LLMs provide for the decoding of the neighborhood environment? 
\end{itemize}
 
To compare the performance of LLMs, we adopt YOLOv11 \cite{yolov11} framework to train a baseline model with six environmental indicators, including streetlight, sidewalk, powerline, apartment, single-lane roadway, and multilane roadway.  In this study, we evaluated four LLMs, including ChatGPT 4o mini \cite{OpenAI_Chat_API}, Gemini 1.5 Pro \cite{gemini}, Claude 3.7 \cite{claudeai}, and Grok-2 \cite{grok2} to study their feasibility to decode the neighborhood environment. 
Our paper is the first study to specifically adopt LLMs to detect environmental indicators using GSV images. We also investigate how various prompting techniques and parameter tuning techniques impact the accuracy of the model at detecting the presence of environmental indicators in the environment. The main contributions of this paper are as follows. 
\begin{itemize}
     \item  We label 1200 Google Street View (GSV) images and train the deep neural network model as a baseline. Our model based on YOLOv11 shows a significant improvement compared to the state-of-the-art. 
    \item We run four LLMs to show the robustness and feasibility of detecting environmental indicators. 
    \item We perform a comprehensive evaluation of the detection of environmental indicators, including the impact of prompting techniques, majority voting, different prompt languages, and parameter fine-tuning.
\end{itemize}

\section{RELATED WORK}
\label{sec:related_work}

\subsection{ML in Identifying Neighborhood Environment}
 In traditional neighborhood environment surveys, researchers would investigate the communities in person to conduct field surveys or send questionnaires \cite{schmidt2014survey, Kang2020UrbanEnvironmentSensing}. ML and DL enable to analyze neighborhood environment through image processing \cite{cheng2018curb} or GIS spatial analysis \cite{pikora2002developing}.  For example, Cheng et al. \cite{cheng2018curb} propose a curb detection method based on support vector machine (SVM) to improve road and sidewalk detection using stereo vision in the urban residential region.
    
    GSV images have become available along with computer vision models to detect the presence of environmental indicators \cite{nguyen2022google, keralis2020health,Nguyen_2021}.
     Keralis et al.\cite{keralis2020health} train a VGG-16 Visual Geometry Group model to detect green space, non-single family homes, single-lane roads and electrical wire.  
     Nguyen et al. \cite{Nguyen_2021} deploy a VGG-19 model to detect street greenness, visible utility wires and dilapidated buildings, and more environmental indicators, including green streets, crosswalks, non-single-family homes, and visible wires are identified in a further study \cite{nguyen2022google}.   
    Alirezaei et al. \cite{alirezaei2023multi} compare ResNet-18 based single task and multitask models to identify dilapidated buildings, chain-link fences, and streetlights to predict environmental indicators that lead to health outcomes such as stroke, arthritis, and high cholesterol. 
    
   However, these models are dependent on a robust set of labeled training data to accurately detect indicators of the built environment. When these models are trained on insufficient data, they may fail to reliably identify environmental indicators. Therefore, it is essential to develop models that can perform effectively even with/without labeled data, ensuring that more complex or subtle effects are captured accurately.
  
    \subsection{LLMs in Image Recognition}
     \textbf{Large Language Models (LLMs).} Recent advances in vision-language models (VLM) have dramatically transformed image recognition tasks by integrating LLMs with computer vision techniques. Pioneering works such as CLIP \cite{radford2021clip}, QwenVL \cite{baidu2023qwen}, and ChatGPT 4o \cite{OpenAI_Chat_API} have demonstrated that by leveraging massive amounts of paired image-text data, models can learn rich, transferable representations that understand both visual cues and their semantic context. The image scene understanding of LLMs could provide a new approach for interpreting environmental indicators such as sidewalks, roadways, traffic scenes, and broader contextual understanding.
    For example,  Zhang et al. \cite{zhang2024good} examine GPT-4v and show that it has remarkable quantitative and qualitative performance in high-level scene understanding, landmark recognition, image captioning, and certain land use classification tasks. The latest open-source Qwen2.5-VL 72B \cite{bai2025qwen2} achieves remarkable results in image understanding and object counting. 

    \textbf{Prompt Engineering.} Prompt engineering significantly impacts the performance of LLMs in downstream tasks.  Numerous extensive studies have studied the identification of the optimal prompt to improve accuracy~\cite{han2023llms,linzbach2023prompt,bareiss2024english}. 
    Linzbach et al.\cite{linzbach2023prompt} investigate BERT variants and show that simple sentence structure performs better for relational knowledge extraction than complex grammatical constructions.
    The study \cite{bareiss2024english} also shows that the English language prompts perform better in natural language inference tasks (e.g., text emotion classification) than the other language prompts.
    To improve image classification accuracy, Han et al.\cite{han2023llms} integrate GPT-4 \cite{OpenAI_Chat_API} and CLIP ViT-B/32 backbone \cite{radford2021clip} to extract the image and text embeddings to iteratively refine the class descriptors with visual feedback from CLIP to guide the optimization process for image classification. 
     
\textbf{Low-shot Image Classification.} Low-shot image classification, including zero-shot and few-shot learning methods learn from a set of class names along with/without a limited set of images.
Recent large-scale pre-trained VLMs like CLIP\cite{radford2021clip} bring training-free zero-shot methods to image classification. ChatGLM \cite{yang2024ChatGLM} aligns the visual and linguistic semantics of CLIP with hierarchical prompts, including global and local prompts to perform image classification, which outperforms unsupervised methods such as WAN \cite{mac2019WAN} and Naive AN \cite{kundu2020NaiveAn}.  
Zhang et al. \cite{zhang2017zsl} use visual space as an embedding space to bridge the gap between seen and unseen categories.
LLaMP \cite{zheng2024lLLaM} uses LLM as a prompt learner for the CLIP model and delivers better performance in both zero-shot generalization and few-shot image classification.

\section{Methodology }
\begin{figure}[htbp]
    \centering
    \includegraphics[width=\columnwidth]{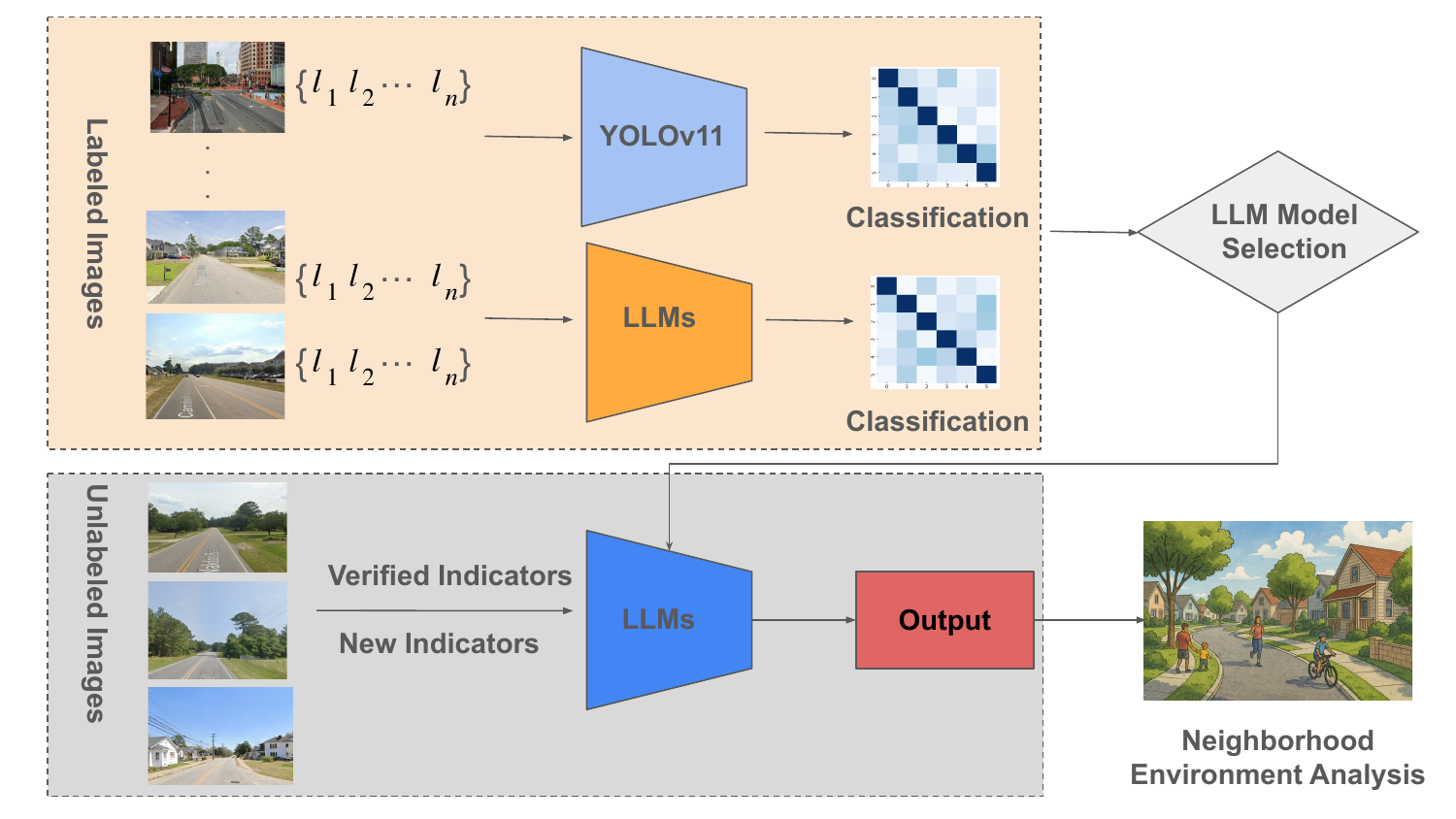} 
    \caption{Overview of Methodology.}
    \label{fig:over_design}
\end{figure}

To address RQ1, we first build a baseline model. Figure~\ref{fig:over_design} illustrates the overview of our proposed methodology. Given images with $n$ labeled environment indicators $\{l_1,l_2,\dots l_n\}$, the supervised learning model (e.g., YOLOv11 for object detection) trains the labeled images and builds the baseline model. The test result will serve as a reference for comparing the performance of LLMs. The LLMs will verify if the indicators exist in the labeled images through prompts.  To address RQ2, we will evaluate several LLMs and extensively examine other factors (e.g., different prompts and languages) which could raise challenges and opportunities for decoding environment indicators. Eventually, we will apply majority voting across three LLMs that have optimum performance to enhance accuracy. Thus, the indicators with higher accuracy can be determined and the LLMs with overall better performance could also take the new indicators to analyze the neighborhood environment. 

\subsection{YOLOv11 Model}
YOLOv11 \cite{yolov11} is the latest iteration in the YOLO (You Only Look Once) series, which brings several key architectural improvements over its predecessors. In essence, YOLOv11 is a highly versatile state-of-the-art computer vision model designed for tasks such as object detection, instance segmentation, image classification, pose estimation, and oriented bounding-box detection.
 YOLOv11 introduces innovative components, such as the C3k2 block, Spatial Pyramid Pooling Fast (SPPF)  module, and Convolutional block with Parallel Spatial Attention (C2PSA) block, which allow it to extract and process image features more effectively than earlier versions, while reducing the number of parameters required. In this study, we use the smallest and fastest YOlOv11 Nano to train the baseline model.

\subsection{LLM Prompt Based Approach}
Research has shown that identifying the optimal prompt is critical to enhance the performance of downstream tasks \cite{han2023llms,linzbach2023prompt}. Therefore, we need to identify an optimal prompt to accurately interpret environmental indicators.

 To achieve this, it is critical to explore various prompt engineering strategies, including zero-shot (e.g., sequential prompt and parallel prompt)
 and parameter tuning, to yield the most reliable output for neighborhood environment assessment. Zero-shot prompting allows models to generate responses without prior image examples. Meanwhile, we also examine the impact of LLM parameter settings and different prompting languages.

\section{Evaluation}
\label{sec:evaluation}
In this section, we describe how we label the GSV images and train the YOLOv11 model. We will conduct ablation experiments to verify the robustness and feasibility of the trained model. Then, we use all the labeled images to conduct extensive experiments in four LLMs, including ChatGPT 4o mini, Gemini 1.5 Pro, Claude 3.7, and Grok-2. 
 
 \subsection{Data Collection}

 We randomly selected 1,200 images from the locations where we segment all roadways with an interval of 50 feet across two counties (e.g., Robeson and Durham counties), covering both rural and urban settings in North Carolina in January 2025. We obtained the coordinates for each location and request images with a resolution of 640×640 pixels from all four directions (e.g., 0 = north, 90 = east, 180 = south, and 270 = west) to fully capture the characteristics of the neighborhood at each coordinate. The GSV image data were accessed lawfully through an API fee. 

 We used the LabelMe tool \cite{russell2008labelme} to label the images. An undergraduate research student manually labeled images, including 1,927 indicator objects. The researcher checked the labels multiple times to ensure consistency and verified the label. The total number of each indicator for streetlight (SL), sidewalk (SW), single-lane road (SR), multilane road (MR), powerline (PL), and apartment (AP) are 206, 444, 346, 505, 301, and 125, respectively. 

\subsection{Baseline Model}
\subsubsection{YOLOv11 model}
 \label{subsub:yolov11}
 We retrained the YOLOv11 Nano model \cite{yolov11} on labeled environment indicators. We randomly split these indicators 70\% as the training set, 20\% as the validation set, and 10\% as the testing set. The samples for each indicator are evenly distributed.
 We trained the model in 20 epochs with a batch size of 16 for the six selected environmental indicators using a Lenovo ThinkPad laptop with a 12 GB Nvidia RTX 3500 GPU. 

 Several metrics are used to calculate the accuracy of an object detection model. The first is the Mean Average Precision (mAP50) at an intersection over union (IoU) with a threshold of 0.50. In this metric, a predicted bounding box is only considered correct if it overlays at least 50\% of a true bounding box. Precision refers to the proportion of correctly identified objects among all the objects a model predicts as positive, while recall measures the proportion of actual positive objects that the model correctly identifies. The F1 score reflects both the precision and the recall of the model. 

 \begin{table}
 \centering
 \caption{Overall Accuracy}
 \label{tab:over_accuracy}
\begin{tabular}{l l l l l} 
 \toprule
 \textbf{Label} & \textbf{Precision} & \textbf{Recall} & \textbf{F1} & \textbf{mAP50}  \\ [0.5ex] 
 \midrule
 Streetlight & 0.993 & 0.995& 0.994& 0.995 \\
 Sidewalk & 1.0 & 0.890 & 0.942 & 0.989 \\
 Single-lane road & 0.938 & 0.871 &0.903& 0.980 \\
 Multilane road & 0.949 & 1.0 &0.974& 0.994 \\
 Powerline & 1.0 & 0.981 &0.990&0.995 \\
 Apartment & 0.954 & 1.0 &0.977& 0.995 \\ [1ex] 
 \midrule
 Average & 0.920 & 0.956 &0.963&0.991  \\ 
 \bottomrule
\end{tabular}
\end{table}

Table \ref{tab:over_accuracy} shows that our model performs very strongly overall, with an average mAP50 of 99.1\% and average F1 score of 96.3\%.,  which generally beats the accuracy of the scene classification models used in previous research \cite{keralis2020health,Nguyen_2021}. The environmental indicators with the highest F1 score is streetlight (0.994), while the lowest F1 score is for single-lane road (0.903).
These results suggest that the object detection model is highly effective for detecting a range of environmental indicators.

\subsubsection{Ablation Study}
We conduct data augmentation and Gaussian Noise to examine the robustness of the trained model. 

\begin{figure}[htbp]
    \centering
    \includegraphics[width=0.6\columnwidth]{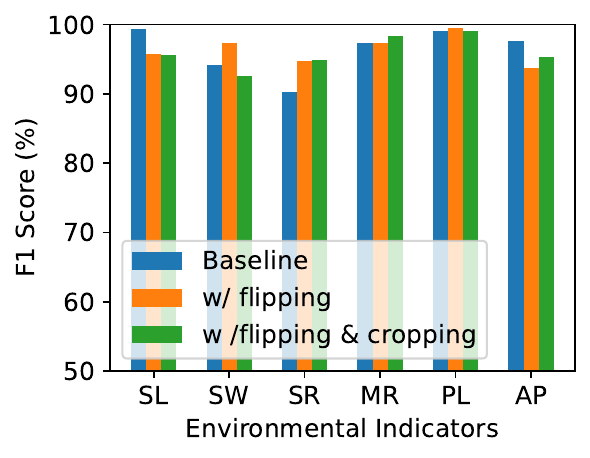} 
    \caption{Accuracy with augmentation}
    \label{fig:augment}
\end{figure}

\paragraphb{Data Augmentation.} We flipped the indicator images in $90^\circ$, $180^\circ$,  and $270^\circ$ to increase the training samples. We follow the same training process as described in Section \ref{subsub:yolov11}. The average F1 score is 96.4\% (mAP50 99.1\%). We use the same approach by adding cropped images, which were randomly cropped by 30\% of the object image area. Combining all images together, the accuracy of F1 is 96\% with the mAP50 of 99\%. As shown in Figure \ref{fig:augment}, data augmentation does not improve overall performance, while making the detection of streetlights and apartments worse. The reason could be that streetlights and apartments have inherent directionality, so that rotating a streetlight to another direction may produce images that would never appear in actual street-level views, confusing the model. Standard augmentations like horizontal flipping and rotation may change the meaning or spatial context of the objects.

\begin{figure}[htbp]
    \centering
    \includegraphics[width=0.5\columnwidth]{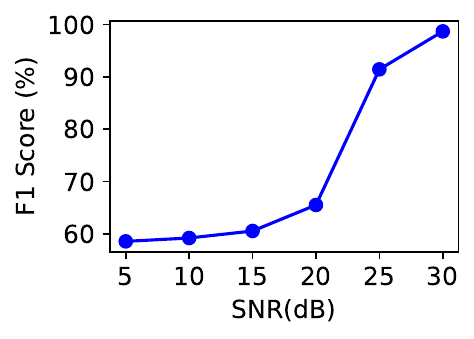} 
    \caption{Impact of different SNR levels}
    \label{fig:snr}
\end{figure}

\textbf{Gaussian Noise.} To evaluate the impact of noise on model performance, we introduce Gaussian noise at Signal-to-Noise Ratio (SNR) levels ranging from 5 to 30 dB with a 5 dB increment. An SNR between 20 and 30 dB represents moderate noise levels, while an SNR between 5 and 20 dB indicates more severe noise that can significantly affect the recognition accuracy.
Figure~\ref{fig:snr} shows that the model maintains an accuracy above 90\% for medium SNR 25 dB and 30 dB, whereas the performance drops to around 60\% at higher noise levels (low SNR). 

\subsubsection{Comparison with Existing Works}
We compare our work to other state-of-the-art indicators detection using GSV. 
 Alirezaei et al. \cite{alirezaei2023multi} compare ResNet-18-based multitask models to identify dilapidated buildings, chain link fences, and streetlights with F1 scores of 95\%, 57\%, and 59\%, respectively. %
 Similarly, Nguyen et al. \cite{nguyen2022google} train a VGG-19 model with accuracies of 88.7\% for street greenness, 97.2\% for crosswalk, 83.0\% for visible utility wires, 82.35\% for non-single family, 88.41\% for single-lane roads, and 83\%  for visible utility wires.  In contrast, our model achieves a significant improvement with a higher average F1 score of 96\%. 

\subsection{LLMs Evaluation}
In this section, we evaluate four LLMs, including ChatGPT 4o mini, Gemini 1.5 Pro, Claude 3.7, and Grok-2, to study their feasibility in detecting indicators of the neighborhood environment. We conduct extensive experiments, including prompt tuning, prompt in different languages, and parameter tuning.
\subsubsection{Prompt Tuning using LLMs}

\begin{table*}[h]
    \centering
    \caption{Result examples of prompts.}
    \label{tab:exampe_table}
    \renewcommand{\arraystretch}{1.5} 
    \resizebox{1.0\textwidth}{!}{
    \begin{tabular}{c c|c c c c c }
        \toprule
        \multirow{7}{*}{\rotatebox{270}{\includegraphics[width=3cm]{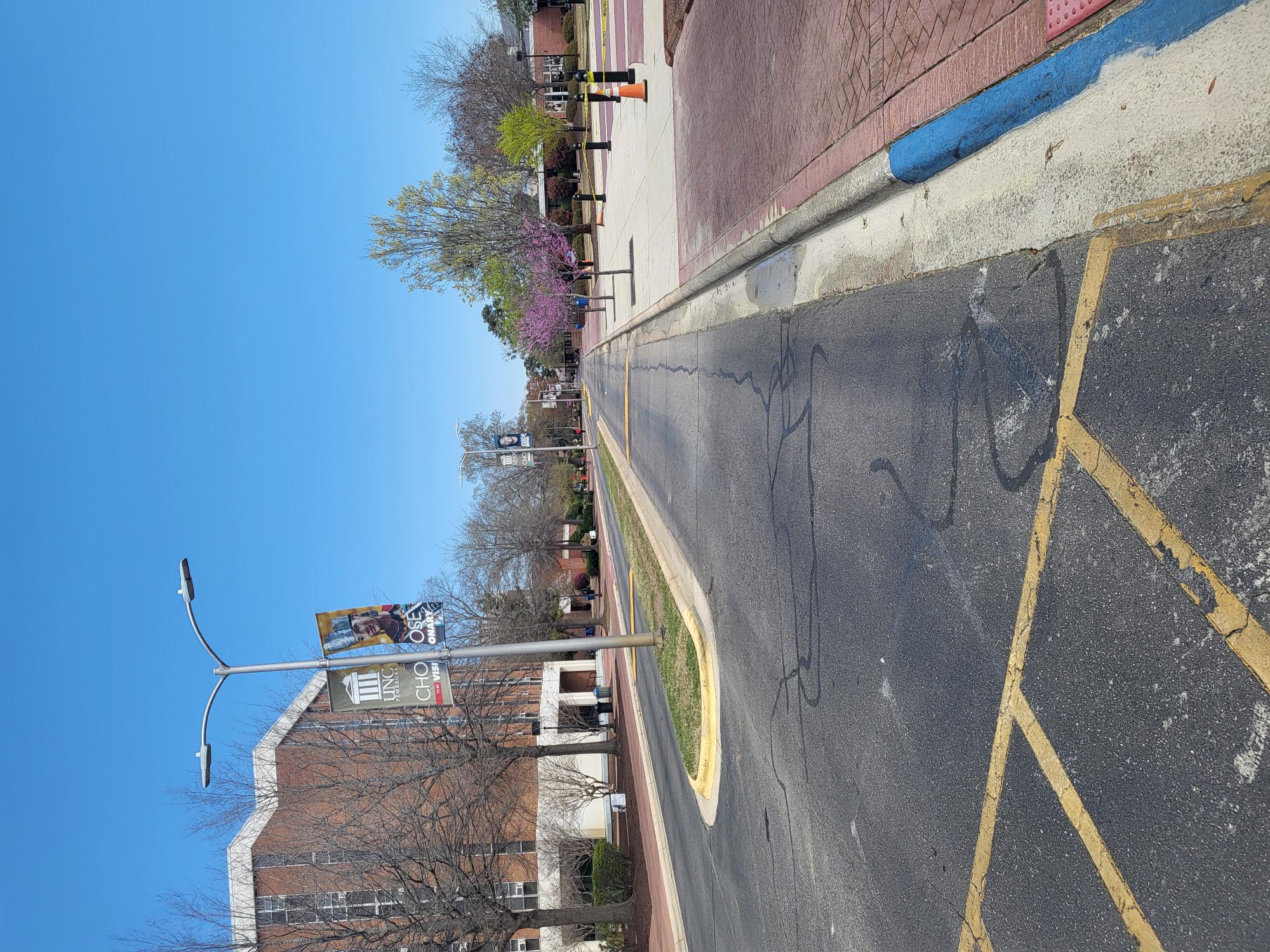}}} & Respond in this format: Yes, No, No, Yes, No, Yes:
 & ChatGPT 4o mini & Gemini1 .5 Pro & Grok 2 & Claude 3.7  \\
        & Is the road shown in the image a multi-lane road (more than one lane per direction)? Respond only with `Yes' or `No'.  & No & No & Yes & No  \\
        & Is the road in the image a single-lane road (one lane per direction)? Respond only with `Yes' or `No'. & Yes & Yes & No & Yes  \\
        & Is there a sidewalk visible in the image? Respond only with `Yes' or `No'.  & Yes & Yes & Yes & Yes \\
        & Is there a streetlight visible in the image? Respond only with `Yes' or `No'.  & Yes & Yes & Yes & Yes  \\
        & Is there a power line visible in the image? Please respond with `Yes' or `No'.  & No & No & No & No \\
        & Is there an apartment visible in the image? Respond only with `Yes' or `No'.  & No & No & Yes & No  \\
        \bottomrule
        
    \end{tabular}
            }
           
\end{table*}

Prompt tuning is a technique used in natural language processing (NLP) to optimize the prompts given to a pre-trained language model, enabling it to generate more accurate or useful responses for a specific task. Table \ref{tab:exampe_table} illustrates an example prompt for recognizing an image with a single lane, streetlight, and sidewalk.

We compare the accuracy for sequential prompting and parallel prompting. 
Sequential Prompting is where the model is asked to detect objects one at a time, often through follow-up prompts. For example, ask if an image contains a sidewalk in one prompt and then ask if it contains a powerline in the next prompt. Parallel prompting is when the model is asked to detect multiple objects in a single-paragraph prompt. For example, ask whether an image contains a sidewalk or a powerline in a single prompt.

We compare the accuracy of ChatGPT and Gemini using sequential prompting, where we asked whether an image contained the environmental indicator in it individually, versus parallel prompting, where in a single prompt we asked it to list all which of the environmental indicators were present. 

We create the prompt for the parallel prompting by concatenating all of the individual sequential prompts and putting "and" in between each one. For example, below is a concatenated parallel prompt. We send a single question prompt to each LLM request. 
\begin{tcolorbox}[colback=blue!5,    
    colframe=black,     
    title= Parallel Prompt,       
    fonttitle=\bfseries,
    coltitle=white,     
    fontupper=\small,   
    boxrule=0.8pt,      
    arc=2pt,            
    left=2pt, right=2pt, top=2pt, bottom=2pt, 
    boxsep=1pt          
    ]
\small
Is the road shown in the image a multi-lane road (more than one lane per direction)? Please answer only with `Yes' or `No'. 

And is the road in the image a single-lane road (one lane per direction)? Please answer only with `Yes' or `No'. 

And is there a sidewalk visible in the image? Respond only with `Yes' or `No'. 

And is there a streetlight visible in the image? Respond only with `Yes' or `No'. 

And is there a powerline visible in the image? Respond only with `Yes' or `No'. 

And is there an apartment visible in the image? Respond only with `Yes' or `No'. 
\end{tcolorbox}

The following are the individual prompts used for the sequential prompts. We combine all single prompts as a paragraph for LLM request. 
\begin{tcolorbox}[colback=blue!5,    
    colframe=black,     
    title=Sequential Prompt,       
    fonttitle=\bfseries,
    coltitle=white,     
    fontupper=\small,   
    boxrule=0.8pt,      
    arc=2pt,            
    left=2pt, right=2pt, top=2pt, bottom=2pt, 
    boxsep=1pt          
    ]
\small
[multilane road prompt] + [single lane road prompt]  + [sidewalk prompt]  + [street light prompt]  + [powerline prompt]  + [apartment prompt]
\end{tcolorbox}

\begin{figure}[htbp]
    \centering
    \subfloat[\small Genimi 1.5 Pro]{%
        \includegraphics[width=0.48\columnwidth]{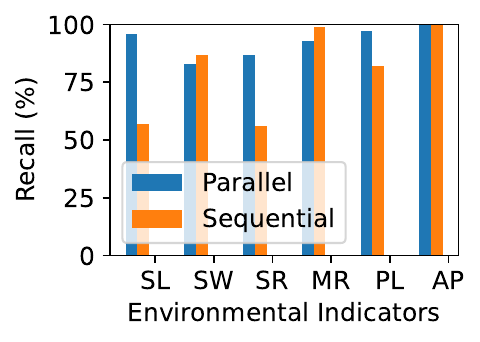}
        \label{fig:genimi}
    }
    \hfill
    \subfloat[\small ChatGPT 4o mini]{%
        \includegraphics[width=0.48\columnwidth]{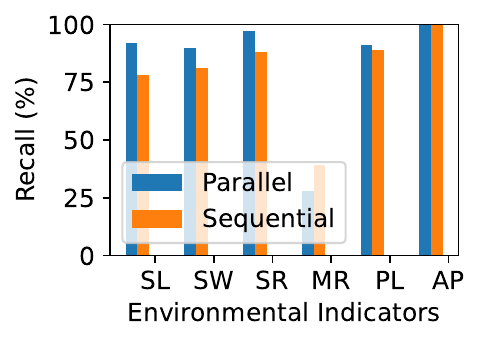}
        \label{fig:chatgpt}
    }
    \caption{Accuracy of LLMs in parallel and sequence prompts.}
    \label{fig:parallel_sequence}
\end{figure}

Figure \ref{fig:parallel_sequence} shows the performance of Genimi 1.5 Pro and ChatGPT 4o mini using parallel and sequential prompts. We compare the recall as known as true positive rate (TPR), which represents how well the model finds all the objects. The result shows that the parallel prompt is with the average recalls of 92\% and 83\% for Genimi and ChatGPT, respectively, while the sequential prompt is with 80\% and 79\%. This suggests that LLMs don't perform well with complex grammatical prompts when using sequential prompts. This finding is also consistent with the result in \cite{linzbach2023prompt} that investigated  BERT variants and showed that simple sentence structure performs better for knowledge extraction than complex grammatical constructions. 

\subsubsection{Majority Voting on LLMs}
\begin{figure}[htbp]
    \centering
    \includegraphics[width=0.6\columnwidth]{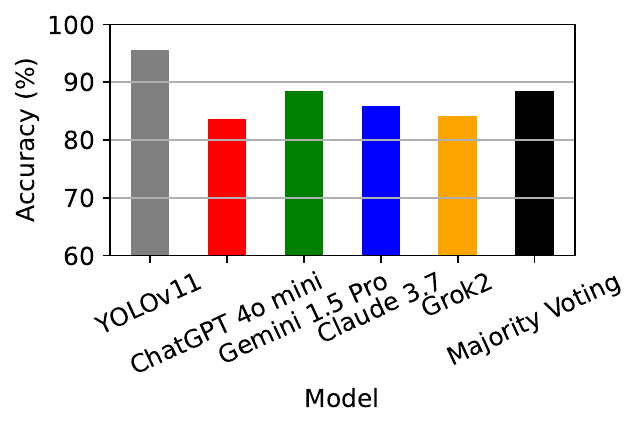} 
    \caption{Accuracy of LLMs and Majority Voting.}
    \label{fig:comparison}
\end{figure}

We also evaluated Claude 3.7 and Grok2 using the same parallel prompt. As shown in Figure~\ref{fig:comparison}, the average accuracies of ChatGPT, Gemini, Claude, and Grok2 are 84\%, 88\%, 86\%, and 84\%, respectively. We list the details of precision, recall, F1 score, and accuracy for each LLM model in Appendix A. 

To improve accuracy, we apply a majority voting scheme on the top three models, including Gemini, Claude, and Grok2, to reach the final prediction when at least two models agree. The accuracies of the majority voting for streetlight, sidewalk, single-lane road, multilane road, powerline, and apartment are 92.86\%, 84.91\%, 68.19\%, 97.07\%, 95.15\%, and 95.15\%, respectively, with an overall average accuracy of 88.5\%. The lower accuracy for single-lane roads (SR) is primarily due to the tendency of LLMs to classify any partial view of a roadway as a single-lane road, regardless of the actual lane count. However, these results demonstrate that majority voting can achieve improved overall accuracy.

\subsubsection{Prompt in Different Languages}

\begin{figure}[htbp]
    \centering
    \includegraphics[width=0.75\columnwidth]{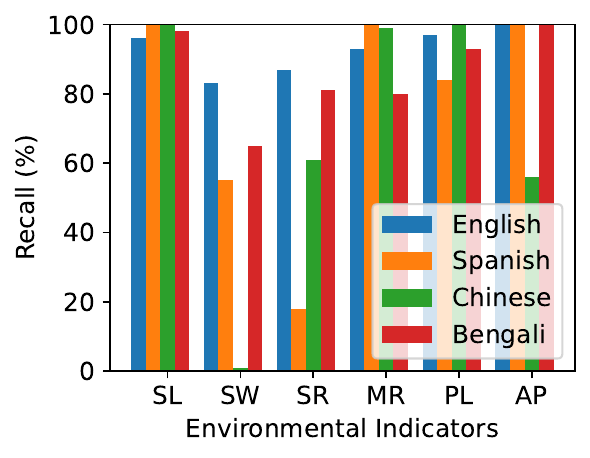} 
    \caption{Accuracy of Different Languages}
    \label{fig:language}
\end{figure}

We evaluate the performance of Gemini 1.5 Pro in multiple languages, including Spanish, simplified Chinese, and Bengali (prompts listed in Appendix B), which represent varying structures and are some of the most spoken languages in the world. We translated the prompts from English to the respective languages with the review of native language speakers to ensure that the meaning was the same.  As shown in Figure \ref{fig:language}, the result shows that English has the best overall recall accuracy of 89.7\%, followed by Bengali with 86\%, Spanish with 76\%, and simplified Chinese with 69\%. This variation is likely influenced by the imbalanced representation of different languages in the multilingual training data of the LLM.

We further analyze the per-class performance. The Chinese-language prompt achieves just 1\% recall for detecting sidewalks (SW), suggesting a severe failure to associate the translated term with relevant visual cues. Similarly, Spanish prompt records only 18\% recall for the identification of single-lane roads (SR), indicating potential ambiguity in the linguistic framework or lack of a robust visual language foundation in that context.

\subsubsection{Parameter Tuning}

    The temperature and Top-P parameters are two parameters that are related to the creativity of the model in its output. Specifically, the temperature regulates the randomness of the model's output by adjusting the probability distribution of the next tokens. A higher temperature (closer to 1 or higher) makes the model more likely to select less probable words, leading to more diverse and creative output. In contrast, a lower temperature (closer to 0) makes the model favor more probable words, resulting in more predictable and focused responses. ChatGPT documentation suggests having a temperature of 0 to .2 for less creative tasks and .8 to 1 for tasks that require more creativity \cite{OpenAI_Chat_API}. We changed the temperature to 0.1 and 1.5 in Gemini testing. We tested all images and the F1 scores are 0.78 and 0.79 for the temperature of 0.1 and 1.5, respectively, which are slightly lower than the default temperature of 1 with the F1 score of 0.81. The finding is aligned with \cite{patel2024exploring}, which examines how changing the temperature could affect the accuracy for clinical tasks, and shows that the accuracy of the model was very close in accuracy for all temperatures. Similarly, another study finds that changing temperature did not have any major impacts on the accuracy of LLMs in handling multiple-choice question tasks \cite{Renze2024}.
    
    The TOP-P parameter specifically controls the maximum range of tokens that the model can consider when generating a word. The model will only consider tokens whose combined probability exceeds this threshold set. A higher TOP-P value (closer to 1) includes a wider range of tokens, promoting diversity in the output. A lower TOP-P value (closer to 0) restricts the model to its most confident predictions, leading to more focused and deterministic outputs. We tested the P-value of 0.5 and 0.75 on Genimi 1.5 Pro. The F1 scores are both 0.79 for the P value of 0.5 and 0.75, respectively, showing slightly lower performance than the default 0.95 (F1 = 0.81)  in Gemini 1.5 Pro. Like temperature, Top-P adjustments mainly influence output variety rather than task performance.

\section{Discussion and Limitation}
This study demonstrates the potential of LLMs as scalable tools for decoding neighborhood environments. There are several limitations and challenges. First, although we have carefully labeled the image data, human error in labeling training data could impact the reliability of the model. 
Second, our experiments reveal that some non-English prompts (e.g., Spanish or Chinese) reduced LLM accuracy by 15–20\% compared to English prompts, which is likely due to uneven multilingual training data in LLMs. This disparity limits the equitable deployment in linguistically diverse regions, where local stakeholders may rely on native languages for analysis. While few-shot learning could partially mitigate this gap, optimizing prompts for cross-lingual robustness and fine-tuning LLMs on a region remains an open challenge.
Third, the success of majority voting across multiple LLMs (e.g., achieving 88\% accuracy) could reduce individual model biases. However, this approach might introduce practical barriers such as computational costs and API latency. 
Lastly, we only use a single-frame image input for each location. In the future, we will incorporate multiple consecutive images in different directions to improve performance, especially for indicators that may be partially occluded in single frames.

\section{Conclusion}
\label{sec:concusion}
In this study, we trained a YOLOv11 model and compared it with the results of the inference of four LLMs. Our results have shown that LLMs perform well in decoding some of the environmental indicators.
We also apply majority voting to achieve an overall accuracy of 88\%, demonstrating that LLM could be a good tool to decode the neighborhood environment without any training effort.

\section*{Acknowledgment}
This work was supported  by the North Carolina Collaboratory HMSI Research Grant Program collab\_486.
\bibliographystyle{IEEEtran}
\bibliography{mybib}

\appendix
\section*{Appendix A: Performance in Other LLMs}
\label{appendix:llms}
Table \ref{tab:chatgpt_accuracy} lists the performance of ChatGPT 4o mini, which achieves a high average recall of 0.91. It can effectively identify true instances across most classes. However, precision varies widely, with significantly low scores for single-lane roads (0.49) and apartments (0.32).
\begin{table}[h]
 \centering
 \caption{Accuracy of ChatGPT 4o mini }
 \label{tab:chatgpt_accuracy}
\begin{tabular}{c| c c c c} 
 \hline
 Label & Precision & Recall & F1 & Accuracy  \\ [0.5ex] 
 \hline
 Streetlight & 0.61& 0.84& 0.70& 0.85\\
 Sidewalk & 0.80& 0.82& 0.81& 0.82\\
 Single-lane road & 0.49& 0.98& 0.66& 0.67 \\
 Multilane road & 0.97& 0.87& 0.92& 0.94 \\
 Powerline & 0.75& 0.94& 0.83& 0.91 \\
 Apartment & 0.32& 1.00& 0.48& 0.84 \\ [1ex] 
 \hline
 Average & 0.66& 0.91& 0.73& 0.84 \\ 
 \hline
\end{tabular}
\end{table}

As shown in Table \ref{tab:genimi_accuracy}, Gemini 1.5 Pro has a similar recall performance of 0.90 but with a higher precision of 0.77 compared to ChatGPT. 
\begin{table}[h]
 \centering
 \caption{Accuracy of Gemini 1.5 Pro}
 \label{tab:genimi_accuracy}
\begin{tabular}{c| c c c c} 
 \hline
 Label & Precision & Recall & F1 & Accuracy  \\ [0.5ex] 
 \hline
 Streetlight & 0.76&0.96&0.85&0.92\\
 Sidewalk & 0.96&0.59&0.73&0.81\\
 Single-lane road & 0.55&0.89&0.68&0.73 \\
 Multilane road & 0.89&0.98&0.93&0.94 \\
 Powerline & 0.91&0.96&0.93&0.97 \\
 Apartment & 0.57&1.00&0.73&0.94 \\ [1ex] 
 \hline
 Average & 0.77&0.90&0.81 &0.88 \\ 
 \hline
\end{tabular}
\end{table}

 A comparative analysis of Grok 2 (shown in Table \ref{tab:grok2_accuracy} and Claude 3.7 (shown in Table \ref{tab:claude_accuracy}) shows that Grok 2 has better performance trade-offs,  which achieves slightly better precision (0.75 vs. 0.72) and F1 score (0.79 vs. 0.78). Grok 2 also shows a significant performance per class in the Powerline (F1 = 0.90 vs. 0.82) and Apartment (F1 = 0.82 vs. 0.70) classification.
\begin{table}[h]
 \centering
 \caption{Accuracy of Grok 2}
 \label{tab:grok2_accuracy}
\begin{tabular}{c| c c c c} 
 \hline
 Label & Precision & Recall & F1 & Accuracy  \\ [0.5ex] 
 \hline
 Street light& 0.76& 0.91& 0.83& 0.91\\
Sidewalk& 0.83& 0.92& 0.88& 0.87\\
Single-lane road& 0.41& 0.99& 0.58& 0.55\\
Multi-lane road& 0.98& 0.56& 0.72& 0.82\\
Powerline& 0.82& 1.00& 0.90& 0.94\\
Apartment& 0.69& 1.00& 0.82& 0.96\\[1ex] 
\hline
Average & 0.75& 0.90& 0.79& 0.84 \\
 \hline
\end{tabular}
\end{table}

\begin{table}[h]
 \centering
 \caption{Accuracy of Claude 3.7}
 \label{tab:claude_accuracy}
\begin{tabular}{c| c c c c} 
 \hline
 Label & Precision & Recall & F1 & Accuracy  \\ [0.5ex] 
 \hline
 Street light& 0.83&0.76&0.79&0.91\\
Sidewalk& 0.76&0.80&0.78&0.80\\
Single-lane road& 0.52&0.99&0.68&0.70\\
Multi-lane road& 0.98&0.85&0.91&0.93\\
Powerline& 0.69&0.99&0.82&0.89\\
Apartment& 0.54&1.00&0.70&0.93\\[1ex] 
\hline
Average & 0.72&0.90&0.78&0.86 \\
 \hline
\end{tabular}
\end{table}


\section*{Appendix B: Prompt in Other Languages}
\label{appendix:languages}
Below are the parallel prompts in the Spanish, Simple Chinese, and Bengali language.
\begin{tcolorbox}[colback=blue!5,    
    colframe=black,     
    title=Spanish Prompt,       
    fonttitle=\bfseries,
    coltitle=white,     
    fontupper=\small,   
    boxrule=0.8pt,      
    arc=2pt,            
    left=2pt, right=2pt, top=2pt, bottom=2pt, 
    boxsep=1pt          
    ]
\small
Por favor, responda exactamente en este formato y ningún otro: sí, no, no, sí, no, no. 

¿La carretera que se muestra en la imagen tiene varios carriles (más de un carril por sentido)? Responda solo con `Sí' o `No'. 

¿La carretera que se muestra en la imagen tiene un solo carril (un carril por sentido)? Responda solo con `Sí' o `No'. 

¿Se ve una acera en la imagen? Responda solo con `Sí' o `No'. 

¿Se ve un alumbrado público en la imagen? Responda solo con `Sí' o `No'. 

¿Se ve un cable eléctrico en la imagen? Responda solo con `Sí' o `No'. 

¿Se ve un apartamento en la imagen? Responda solo con `Sí' o `No'. 
\end{tcolorbox}

\begin{figure}[htbp]
    \centering
    \includegraphics[width=1.01\columnwidth]{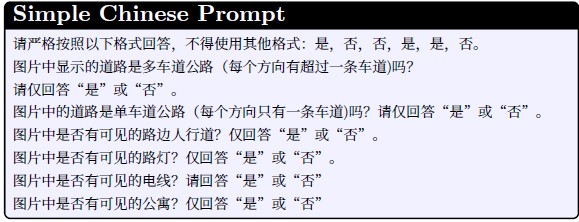} 
\end{figure}

\begin{figure}[!t]
    \centering
    \includegraphics[width=\columnwidth]{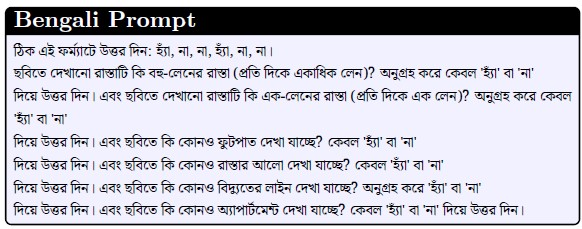} 
\end{figure}

\end{document}